\documentclass[11pt]{article}

\usepackage[preprint]{acl}

\usepackage{times}
\usepackage{latexsym}

\usepackage[T1]{fontenc}

\usepackage[utf8]{inputenc}

\usepackage{microtype}
\usepackage{inconsolata}
\usepackage{graphicx}
\usepackage{amsmath}
\usepackage{amssymb}
\usepackage{booktabs}
\usepackage{multirow}
\usepackage[table]{xcolor}
\usepackage{adjustbox}
\usepackage{hyperref}
\usepackage{enumitem}
\usepackage{fontawesome5}
\usepackage{xcolor}
\usepackage{subcaption}
\definecolor{githubblue}{HTML}{0366D6}

\hypersetup{
    colorlinks=true,
    urlcolor=githubblue,
}
\newcommand{\secref}[1]{\hyperref[#1]{\S\ref*{#1}}}

\definecolor{avgshade}{HTML}{72B879}
\definecolor{passshade}{HTML}{6FA8DC}
\definecolor{lenshade}{HTML}{9D7AC2}
\usepackage{inconsolata}

\usepackage{graphicx}

\title{PowerOPD: Stabilizing On-Policy Distillation with Bounded Power Transformation}

\author{
  \textbf{Anhao Zhao\textsuperscript{1,2},
  Junlong Tong\textsuperscript{1,3},
  Yingqi Fan\textsuperscript{1},
  Ping Nie\textsuperscript{4}},
  \textbf{Wenjie Li\textsuperscript{2},
  Xiaoyu Shen\textsuperscript{1}\thanks{Corresponding Author}}
  \\
  \textsuperscript{1} Eastern Institute of Technology, Ningbo
  \quad
  \textsuperscript{2}The Hong Kong Polytechnic University
  \\
  \textsuperscript{3}Shanghai Jiao Tong University
  \quad
  \textsuperscript{4}University of Waterloo
  \\
  {\texttt{anhao.zhao@connect.polyu.hk}\quad
  \texttt{xyshen@eitech.edu.cn}}
}

\begin{document}
\maketitle
\begin{abstract}
Standard on-policy distillation (OPD) for large language models estimates the reverse-KL objective using student-sampled tokens, yielding an unbiased single-sample Monte Carlo estimator that avoids vocabulary-wide computation. However, we show that this estimator suffers from severe training pathologies in practice: sample inefficiency, unstable generation dynamics, and a substantial performance gap compared to exact full-vocabulary OPD.
Reward-level diagnosis traces these pathologies to the log-ratio reward, which is unbounded by construction, producing extremely high-variance gradients concentrated at early positions and persisting throughout training; standard post-hoc scaling fail as they operate only after this distortion occurs.
To solve this problem, we propose \emph{PowerOPD}: a family of natively bounded, sign-consistent rewards from the Box-Cox power transformation, parameterized by $\alpha > 0$, of which the log-ratio is the degenerate $\alpha \to 0$ limit.
Across six mathematical reasoning benchmarks and four Qwen3 teacher--student pairs, PowerOPD achieves benchmark-averaged Avg@8/Pass@8 gains of up to $\mathbf{+6.37/+5.71}$ over vanilla OPD, $\mathbf{+3.01/+3.54}$ over post-hoc stabilization, and $\mathbf{+2.59/+8.90}$ over full-vocabulary OPD, while reducing wall-clock time by $59.2\%$ and peak GPU memory by $23.1\%$.
Larger $\alpha$ generally improves accuracy, consistently shortens responses, and keeps gradient norms more than $\mathbf{3{,}000\times}$ smaller than vanilla OPD.
We release our code at \textcolor{githubblue}{\faGithub}\ \href{https://github.com/EIT-NLP/PowerOPD}{\textbf{\texttt{EIT-NLP/PowerOPD}}}.
\end{abstract}

\section{Introduction}
\label{sec:intro}

On-policy distillation (OPD) has rapidly become a standard component of LLM post-training \citep{gu2024minillm,agarwal2024gkd,song2026survey}. By grounding supervision in student-generated trajectories, OPD mitigates the exposure bias of supervised fine-tuning (SFT) and classical off-policy distillation \citep{bengio2015scheduled,hinton2015distilling,deepseekr1}, while providing dense token-level feedback compared to sparse-reward reinforcement learning (RL). These advantages have made OPD a widely adopted bridge between SFT and RL in modern post-training~\cite{qwen3,glm5,deepseekv4,mimo_v2_flash}.

\begin{figure}[t]
  \centering
  \includegraphics[width=\columnwidth]{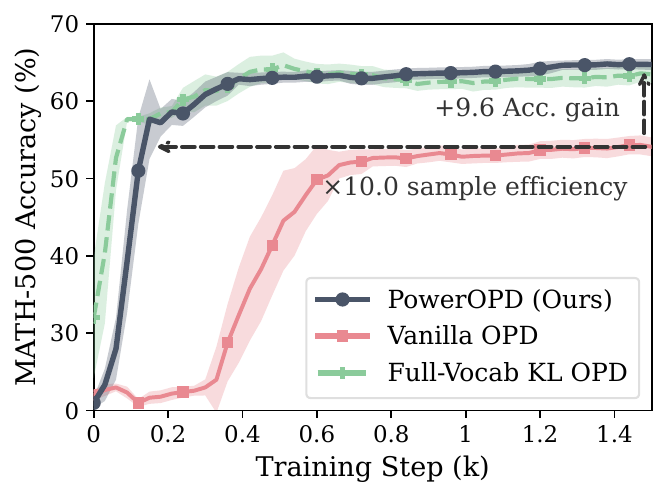}
  \vspace{-20pt}
\caption{PowerOPD achieves $\mathbf{+9.6}$ accuracy gain and $\mathbf{10\times}$ sample efficiency over vanilla OPD, matching or exceeding full-vocabulary KL OPD at 59.2\% less wall-clock time (Qwen3-1.7B $\leftarrow$ Qwen3-4B, \textsc{MATH-500}).}
    \vspace{-5pt}
  \label{fig:first_page}
\end{figure}

In its original form, OPD minimizes the reverse-KL divergence between teacher and student distributions by comparing probabilities over the full vocabulary at each generation step. However, computing the full-vocabulary objective is expensive in practice. Consequently, modern implementations estimate the objective using student-sampled tokens, yielding an unbiased single-sample Monte Carlo estimator \citep{kevin2025opd,jin2026entropy,ko2026reopold,jia2026asymmetric,liu2026teacherguidedpolicyoptimizationonpolicy,zhao2026decouplingkltrajectoriesunified,zhao2026opdrethinkingadvantagedesign}\footnote{As this sampled-token formulation has become the de facto implementation of OPD due to its substantially lower cost, we refer to it as \emph{vanilla OPD} throughout this paper.}.

Despite its widespread adoption, we observe that vanilla OPD exhibits severe \emph{pathological training dynamics} in practice. In a representative Qwen3-4B teacher and Qwen3-1.7B-Base student setting on \textsc{MATH-500}, validation accuracy initially decreases and does not recover for hundreds of training steps despite dense token-level supervision, indicating sample inefficiency. Meanwhile, response length undergoes large oscillations before stabilizing, indicating that the student repeatedly enters unstable generation regimes. Even after convergence, vanilla OPD reaches only 54.93\% accuracy, trailing full-vocabulary OPD by 8.19 points. Together, these failures suggest that \emph{the Monte Carlo approximation underlying vanilla OPD introduces optimization difficulties that significantly limit both training efficiency and final performance}.

To understand the source of these pathological training dynamics, we examine the token-level reward that directly weights each policy-gradient update: the teacher-student log-probability ratio. Our diagnosis shows that the unbounded log-ratio reward distorts the update signal along three dimensions:
(i) extreme reward variance, where reward values plummet to nearly $-50$, allowing single rare tokens to dominate gradient updates and trigger generative instability; (ii) early-position extremes, where massive reward magnitudes disproportionately hit the early positions of student rollouts, destabilizing the prefix distribution and causing cascading errors that drive sample inefficiency; and (iii) persistent extreme rewards, where these massive positive and negative values fail to decay, injecting instability throughout the entire optimization process. Notably, applying standard RL reward-stabilization tools \citep{mnih2015human,schulman2017ppo}, such as clipping, tanh compression, and z-score normalization, does not resolve these issues, indicating that \textit{the instability originates from the unbounded log-ratio reward itself rather than from insufficient post-hoc scaling}.

Recognizing that unboundedness is the root cause, we reframe OPD reward design as learning a principled probability-to-reward mapping. A well-conditioned OPD reward must satisfy two properties: boundedness, to prevent rare Monte Carlo events from inducing catastrophic gradient updates, and sign consistency, to ensure the reward sign correctly aligns with the teacher–student probability gap (i.e., yielding a positive reward when the teacher assigns a higher probability than the student, and vice versa). We retain the transform-then-subtract structure of the standard reward, \(h(\pi_T)-h(\pi_\theta)\) because it guarantees sign consistency for any strictly increasing \(h\). To achieve boundedness without losing this directional signal, we instantiate \(h\) using the Box–Cox power family \citep{boxcox1964}. This yields \textbf{PowerOPD}: a family of bounded and sign-consistent OPD rewards parameterized by \(\alpha>0\). While the unstable standard log-ratio reward represents the degenerate \(\alpha\to0\) limit, our formulation ensures the reward remains strictly bounded for any \(\alpha>0\).

We evaluate PowerOPD on six mathematical reasoning benchmarks across four teacher--student pairs from the Qwen3 family (0.6B and 1.7B students; 4B and 8B teachers).
As shown in Figure~\ref{fig:first_page}, using a Qwen3-1.7B-Base student and Qwen3-4B teacher, PowerOPD achieves a $\mathbf{+9.6}$ accuracy gain over vanilla OPD and reaches the same accuracy level with $\mathbf{10\times}$ fewer training steps.
Across the full benchmark evaluation, PowerOPD achieves benchmark-averaged Avg@8/Pass@8 gains of up to $\mathbf{+6.37/+5.71}$ over vanilla OPD, $\mathbf{+3.01/+3.54}$ over post-hoc stabilization, and $\mathbf{+2.59/+8.90}$ over full-vocabulary OPD, with individual-benchmark gains reaching $\mathbf{+16.75/+15.00}$, $\mathbf{+8.43/+7.50}$, and $\mathbf{+11.60/+25.00}$, respectively, while reducing wall-clock time per step by $59.2\%$ and peak GPU memory by $23.1\%$ relative to full-vocabulary OPD.
Notably, PowerOPD scales with $\alpha$: larger $\alpha$ generally improves accuracy, shortens responses, and stabilizes training dynamics.
We further show that this scalability is mechanistically grounded: larger $\alpha$ suppresses rewards for tokens that both models assign low probability, while focusing learning on tokens that either the teacher or student considers likely.
Finally, gradient tracking shows that PowerOPD keeps norms more than $\mathbf{3{,}000\times}$ below vanilla OPD's initial spike, while post-hoc methods only partially stabilize training.

\begin{figure*}[t]
  \centering
  \begin{minipage}[t]{0.32\textwidth}
    \centering
    \includegraphics[width=\linewidth]{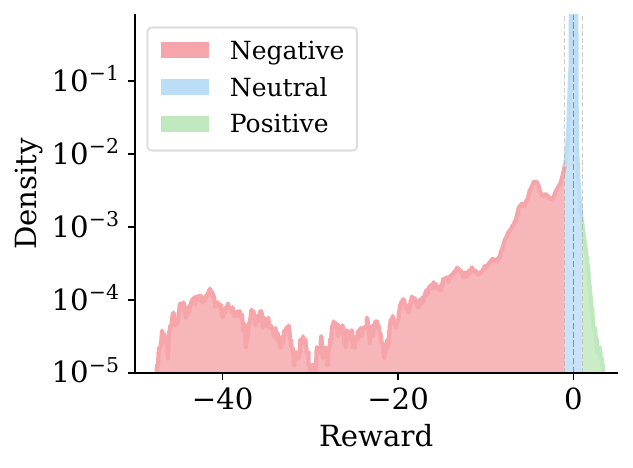}\\[-0.5ex]
    \textbf{(a)}
  \end{minipage}\hfill
  \begin{minipage}[t]{0.32\textwidth}
    \centering
    \includegraphics[width=\linewidth]{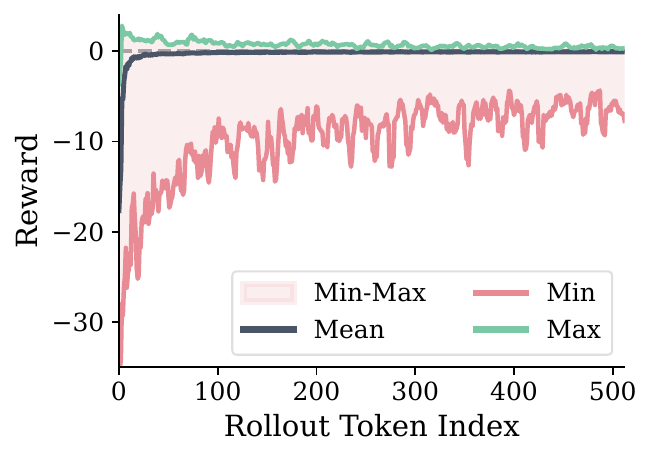}\\[-0.5ex]
    \textbf{(b)}
  \end{minipage}\hfill
  \begin{minipage}[t]{0.32\textwidth}
    \centering
    \includegraphics[width=\linewidth]{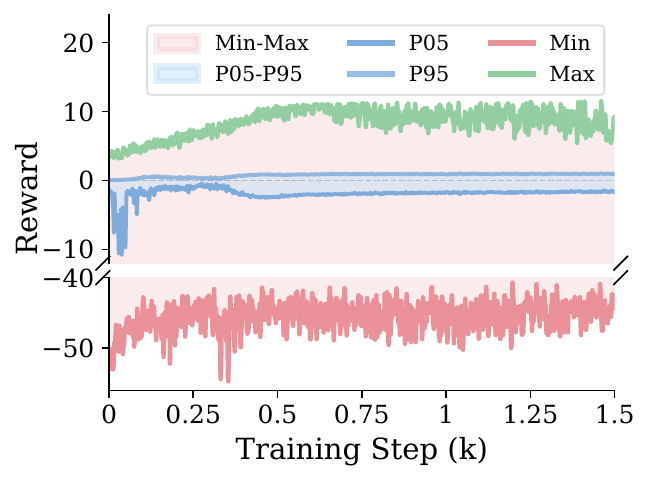}\\[-0.5ex]
    \textbf{(c)}
  \end{minipage}
  \vspace{-8pt}
    \caption{
    Pathological OPD rewards. The OPD reward shows (a) high variance with a heavy negative tail, (b) early-position extreme values, and (c) persistent extremes throughout training.
    }
    \vspace{-12pt}

  \label{fig:reward-diagnosis}
\end{figure*}

\section{Preliminaries}
\label{sec:Preliminaries}

\subsection{On-Policy Distillation}
\label{sec:opd}

On-policy distillation (OPD)~\citep{agarwal2024gkd,gu2024minillm,song2026survey} has emerged as a standard stage in LLM post-training pipelines~\citep{qwen3, HY_MT1.5, glm5, nemotron2, deepseekv4, HY_Embodied_0.5, mimo_v2_flash, kat_coder_v2, Qwen3.5_Omni}. It trains a student policy $\pi_\theta$ to match a stronger teacher policy $\pi_T$ on trajectories generated by the student itself. This on-policy training reduces exposure bias by aligning the training-time contexts with the student's inference-time generation. OPD minimizes the reverse KL divergence, which encourages mode-seeking toward the teacher's dominant modes~\citep{kevin2025opd}:
\[
D_{\mathrm{KL}}(\pi_\theta \| \pi_T)
=
\mathbb{E}_{x \sim \mathcal{D},\, o \sim \pi_\theta(\cdot \mid x)}
\left[
\log
\frac{
\pi_\theta(o \mid x)
}{
\pi_T(o \mid x)
}
\right],
\]
where \(x\) is a prompt from \(\mathcal{D}\) and \(o=(o_1,\ldots,o_{|o|})\) is a student-generated response.

\subsection{OPD as Dense-Reward RL}
\label{sec:opd-as-rl}

Using the autoregressive factorization of $\pi_\theta$ and $\pi_T$, OPD maximizes the negative reverse-KL objective in the token-level form
\[
J_{\mathrm{OPD}}(\theta)
=
\mathbb{E}_{x \sim \mathcal{D},\, o \sim \pi_\theta(\cdot \mid x)}
\left[
\sum_{t=1}^{|o|}
\log
\frac{
\pi_T(o_t \mid c_t)
}{
\pi_\theta(o_t \mid c_t)
}
\right],
\]
where $c_t=(x,o_{<t})$ denotes the context before token $o_t$. Following recent practice~\citep{kevin2025opd, ko2026reopold, oh2026kl}, this objective is optimized as policy-gradient RL~\citep{williams1992reinforce, sutton1999policygrad}:
\[
\begin{aligned}
&\nabla_\theta J_{\mathrm{OPD}}(\theta)
=
\mathbb{E}_{x \sim \mathcal{D},\, o \sim \pi_\theta(\cdot \mid x)}\\
&\Bigg[
\sum_{t=1}^{|o|}
\log
\frac{\pi_T(o_t \mid c_t)}
{\pi_\theta(o_t \mid c_t)}
\nabla_\theta \log \pi_\theta(o_t \mid c_t)
\Bigg].
\end{aligned}
\]
Appendix~\ref{app:opd-pg-derivation} provides the detailed derivation. This policy-gradient form identifies the stop-gradient log-ratio term as the OPD token-level reward:
\begin{equation}
r_t^{\mathrm{OPD}}(c_t,o_t)
=
\log
\frac{
\pi_T(o_t \mid c_t)
}{
\pi_\theta(o_t \mid c_t)
}.
\label{eq:opd-token-reward}
\end{equation}
Unlike the sparse sequence-level rewards used in outcome-based RL~\citep{deepseekmath, deepseekr1}, the OPD reward is dense and depends explicitly on the current student policy $\pi_\theta(o_t \mid c_t)$.
\section{Empirical Failure Modes of OPD}
\label{sec:failure-modes}

We first identify pathological OPD training dynamics: sample inefficiency, unstable generation behavior, and a persistent gap to full-vocabulary OPD (\secref{subsec:training-dynamics}). We then trace them to high-variance log-ratio rewards, whose extremes concentrate early and persist throughout training (\secref{subsec:high-variance}). Finally, we show that standard RL reward-stabilization strategies fail to fix these pathologies (\secref{subsec:standard-fixes}).

\subsection{Pathological Training Dynamics}
\label{subsec:training-dynamics}

\begin{figure}[t]
  \centering
  \begin{minipage}[t]{0.49\columnwidth}
    \centering
    \includegraphics[width=\linewidth]{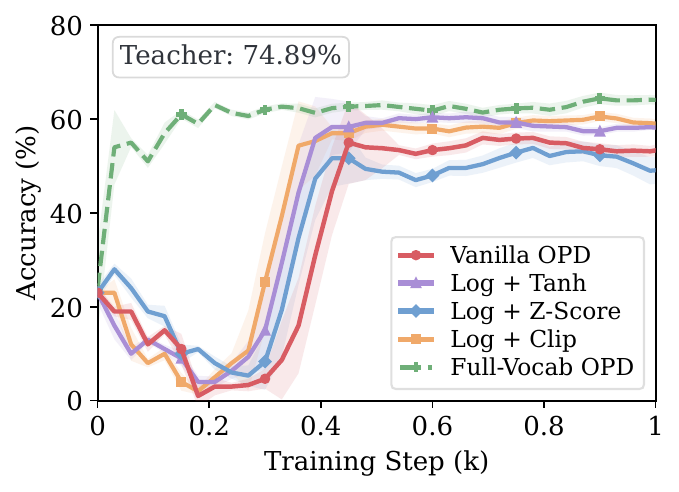}\\[-0.5ex]
    \textbf{(a)}
  \end{minipage}\hfill
  \begin{minipage}[t]{0.49\columnwidth}
    \centering
    \includegraphics[width=\linewidth]{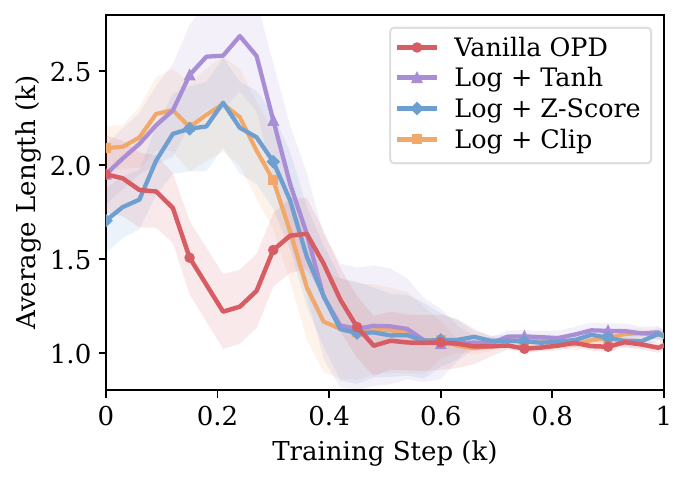}\\[-0.5ex]
    \textbf{(b)}
  \end{minipage}
  \vspace{-7pt}
    \caption{
    Pathological OPD training dynamics. 
    OPD shows (a) delayed, unstable accuracy far below full-vocabulary OPD and (b) response-length oscillations.
    }
    \vspace{-15pt}
  \label{fig:standard-fixes}
\end{figure}

We begin by empirically examining the training dynamics of OPD. We train a Qwen3-1.7B student with a Qwen3-4B teacher and monitor performance on \textsc{MATH-500}  \citep{hendrycks2021math}. As shown in \autoref{fig:standard-fixes}, OPD exhibits three pathological behaviors. \emph{(i) OPD is sample-inefficient.} Accuracy initially decreases and does not begin to recover until roughly 300 training steps, despite the availability of dense token-level supervision. \emph{(ii) OPD is unstable in both accuracy and generation behavior.} Validation accuracy fluctuates substantially, while the average validation response length undergoes large oscillations and stabilizes only after around 400 steps, suggesting that the student policy moves through unstable generation regimes during training. \emph{(iii) OPD converges to a substantially weaker policy than full-vocabulary OPD.}
Full-vocabulary OPD computes the distillation signal over the entire vocabulary rather than only the sampled student token \citep{zhao2026decouplingkltrajectoriesunified}. On \textsc{MATH-500}, OPD plateaus at 54.93\%, whereas full-vocabulary OPD reaches 63.12\%, leaving an 8.19-point gap and recovering only 69.44\% of the teacher's validation accuracy.

\subsection{Pathological Reward Distributions}
\label{subsec:high-variance}

To understand what undermines OPD training, we examine its dense token-level rewards. Since these rewards directly scale the policy-gradient updates, their distribution determines how stable and reliable the optimization signal is.

\paragraph{OPD rewards exhibit extremely high variance.}
We examine the reward distribution induced by OPD before training. Using the Qwen3-1.7B-Base student and the Qwen3-4B teacher, we sample 512 examples from \textsc{MATH-500}, generate student rollouts, and compute the OPD reward \(r_t^{\mathrm{OPD}}(c_t,o_t)\) for every rollout token. We collect these token-level rewards across all rollouts and plot their empirical distribution. As shown in \autoref{fig:reward-diagnosis}(a), OPD rewards span an extremely wide range, with the negative tail reaching nearly \(-50\). This high-variance distribution is a direct consequence of the log difference, which can amplify teacher--student probability discrepancies into unbounded reward magnitudes \citep{ko2026reopold, jia2026asymmetric}. Because each policy-gradient update is directly scaled by this reward scalar, extreme OPD rewards artificially inflate gradient variance and make optimization fragile.

\paragraph{Extreme rewards concentrate at early rollout positions.}
We group OPD rewards by rollout token index and compute position-wise mean, minimum, and maximum values, filtering positions with too few samples. As shown in \autoref{fig:reward-diagnosis}(b), the most extreme rewards occur near the beginning of student rollouts. This pattern is especially harmful in on-policy training: unstable rewards on early tokens can shift the prefix distribution, and subsequent rollouts then condition on these shifted prefixes, propagating instability to later tokens \citep{ liu2026prefixteachsuffixfade}. This feedback loop destabilizes the training distribution and contributes to fluctuations in both validation accuracy and response length.
\paragraph{Extreme rewards persist across training.}
We track the reward dynamics during OPD training by recording batch-level statistics over all rollout-token rewards at each training step, including the minimum, maximum, and the \(5\)th--\(95\)th percentile range. As shown in \autoref{fig:reward-diagnosis}(c), extreme positive and negative rewards persist throughout training and remain close to their initial scale. This indicates that high-variance rewards are not merely an initialization artifact or an early-stage transient. Instead, OPD is exposed to severe token-level rewards and penalties throughout optimization.

\subsection{Post-Hoc Reward Stabilization Fails}
\label{subsec:standard-fixes}

\begin{table}[t]
  \centering
  \small
  \begin{tabular}{lll}
    \toprule
    \textbf{Method} & \textbf{Transformation} & \textbf{Effect} \\
    \midrule
    \emph{Clip} 
      & $\mathrm{clip}(r_t^{\mathrm{OPD}}, c_l, c_h)$ 
      & Hard truncation \\
    \emph{Tanh} 
      & $\tanh(r_t^{\mathrm{OPD}} / \tau)$ 
      & Smooth compression \\
    \emph{Z-Score} 
      & $(r_t^{\mathrm{OPD}} - \mu_B) / \sigma_B$ 
      & Batch normalization \\
    \bottomrule
  \end{tabular}
  \vspace{-7pt}
    \caption{Post-hoc OPD reward transformations.}
    \vspace{-15pt}
  \label{tab:standard-fixes}
\end{table}

\begin{table*}[t]
\centering

\begin{subtable}{\linewidth}
\centering
\scriptsize
\setlength{\tabcolsep}{1.0pt}
\renewcommand{\arraystretch}{1.05}
\caption{Teacher: Qwen3-4B $\to$ Student: Qwen3-0.6B-Base}
\label{tab:main-4b}
\begin{adjustbox}{max width=\linewidth}
\begin{tabular}{@{}ll lll@{\hskip 6pt}lll@{\hskip 6pt}lll@{\hskip 6pt}lll@{\hskip 6pt}lll@{\hskip 6pt}lll@{\hskip 6pt}lll@{\hskip 6pt}lll@{}}
\toprule
 &  & \multicolumn{3}{c}{GSM8K} & \multicolumn{3}{c}{MATH500} & \multicolumn{3}{c}{AMC23} & \multicolumn{3}{c}{AIME24} & \multicolumn{3}{c}{AIME25} & \multicolumn{3}{c}{Minerva} & \multicolumn{3}{c}{Olympiad} & \multicolumn{3}{c}{Mean} \\
Class & Param & Avg & Pass & Len & Avg & Pass & Len & Avg & Pass & Len & Avg & Pass & Len & Avg & Pass & Len & Avg & Pass & Len & Avg & Pass & Len & Avg & Pass & Len \\
\cmidrule(lr){3-5} \cmidrule(lr){6-8} \cmidrule(lr){9-11} \cmidrule(lr){12-14} \cmidrule(lr){15-17} \cmidrule(lr){18-20} \cmidrule(lr){21-23} \cmidrule(lr){24-26}
\midrule
\multirow{1}{*}{Full-vocab} & -- & 66.04 & 87.08 & 4087 & 43.90 & 54.84 & 4096 & 28.75 & 40.00 & 4096 & \textbf{3.33} & \textbf{10.00} & 4096 & 1.67 & 3.33 & 4096 & 11.03 & 15.81 & 4077 & 5.62 & 25.00 & 4094 & 22.91 & 33.72 & 4092 \\
\cmidrule(lr){1-26}
\multirow{4}{*}{Vanilla} & -- & 61.50 & 86.81 & 4078 & 36.02 & 61.20 & 4084 & 14.37 & 47.50 & 4096 & 1.67 & \underline{6.67} & 4096 & 0.83 & 3.33 & 4085 & 6.99 & 17.65 & 4065 & 12.50 & 30.07 & 4087 & 19.13 & 36.18 & 4084 \\
 & + Clip & 65.90 & 87.26 & 498 & 42.27 & 63.24 & 1483 & 21.88 & 52.50 & 2669 & 0.42 & 3.33 & 3722 & 1.25 & 6.67 & 3842 & 10.06 & 21.69 & 1378 & 15.67 & 33.78 & 2809 & 22.49 & 38.35 & 2343 \\
 & + Tanh & 65.38 & 87.00 & 502 & 43.88 & 62.00 & 1493 & 22.69 & 52.50 & 2609 & 0.42 & 3.33 & 3786 & 0.42 & 3.33 & 3880 & 10.03 & 23.00 & 1329 & 4.62 & 15.00 & 3872 & 21.06 & 35.17 & 2496 \\
 & + Z-Score & 56.21 & 86.43 & 437 & 35.75 & 62.26 & 1571 & 14.37 & 45.00 & 2648 & 0.42 & 3.33 & 3691 & 0.42 & 3.33 & 3703 & 8.00 & 20.22 & 1410 & 12.61 & 31.41 & 2750 & 18.25 & 36.00 & 2316 \\
\cmidrule(lr){1-26}
\multirow{8}{*}{PowerOPD} & $\alpha=0.1$ & 68.58 & \underline{89.23} & 404 & 43.62 & 64.92 & 1482 & 28.12 & \underline{57.50} & 2558 & 2.50 & \underline{6.67} & 3623 & 0.42 & 3.33 & 3806 & 9.38 & 20.96 & 1391 & 16.48 & 34.67 & 2776 & \cellcolor{avgshade!12}24.16 & \cellcolor{passshade!12}39.61 & \cellcolor{lenshade!12}2291 \\
 & $\alpha=0.5$ & 68.67 & \underline{89.23} & 407 & 45.29 & \textbf{66.92} & 1438 & 26.88 & 55.00 & 2617 & 2.50 & \underline{6.67} & 3633 & 1.25 & 6.67 & 3662 & 10.62 & 23.16 & 1225 & 16.96 & \underline{36.15} & 2704 & \cellcolor{avgshade!33}24.60 & \cellcolor{passshade!38}40.54 & \cellcolor{lenshade!30}2241 \\
 & $\alpha=1$ & \underline{69.48} & \textbf{90.22} & 396 & 44.34 & 66.18 & 1451 & 24.38 & 47.50 & 2628 & 2.08 & 3.33 & 3583 & \underline{2.50} & \underline{10.00} & 3703 & 10.57 & \textbf{24.26} & 1236 & \textbf{17.22} & 35.85 & 2701 & \cellcolor{avgshade!22}24.37 & \cellcolor{passshade!12}39.62 & \cellcolor{lenshade!21}2243 \\
 & $\alpha=5$ & 69.19 & 88.55 & \textbf{374} & \textbf{45.66} & \underline{66.38} & 1417 & \textbf{31.12} & \underline{57.50} & 2456 & 2.50 & 3.33 & 3610 & 0.83 & 6.67 & 3629 & \textbf{12.22} & 23.63 & \underline{1129} & 16.87 & 35.85 & 2677 & \cellcolor{avgshade!74}\underline{25.48} & \cellcolor{passshade!30}40.27 & \cellcolor{lenshade!39}2185 \\
 & $\alpha=10$ & 69.42 & 88.55 & 393 & \underline{45.54} & 65.92 & 1388 & \underline{29.69} & \underline{57.50} & 2426 & 2.08 & 3.33 & 3597 & 1.25 & 3.33 & 3612 & 11.31 & \underline{23.90} & \textbf{1124} & 17.11 & 35.26 & 2664 & \cellcolor{avgshade!61}25.20 & \cellcolor{passshade!14}39.68 & \cellcolor{lenshade!48}2172 \\
 & $\alpha=50$ & 69.25 & 87.11 & \underline{385} & 45.12 & 66.00 & 1367 & 24.69 & \textbf{60.00} & 2358 & \underline{2.92} & \textbf{10.00} & 3283 & 0.83 & 6.67 & 3502 & 11.31 & 22.43 & 1149 & \underline{17.20} & \textbf{36.59} & 2571 & \cellcolor{avgshade!27}24.47 & \cellcolor{passshade!58}\underline{41.26} & \cellcolor{lenshade!57}2088 \\
 & $\alpha=100$ & 68.70 & 86.88 & \underline{385} & 44.57 & 65.66 & \underline{1363} & 28.12 & 57.00 & \underline{2230} & 2.50 & \textbf{10.00} & \underline{3282} & 0.83 & 6.67 & \underline{3324} & 10.75 & 23.69 & 1225 & 17.04 & \underline{36.15} & \underline{2490} & \cellcolor{avgshade!35}24.64 & \cellcolor{passshade!47}40.86 & \cellcolor{lenshade!66}\underline{2043} \\
 & $\alpha=500$ & \textbf{69.77} & 87.49 & 397 & 44.62 & 65.50 & \textbf{1324} & 29.25 & \underline{57.50} & \textbf{2049} & \textbf{3.33} & \textbf{10.00} & \textbf{3077} & \textbf{2.92} & \textbf{13.33} & \textbf{2819} & \underline{11.40} & 23.53 & 1273 & \textbf{17.22} & 35.85 & \textbf{2185} & \cellcolor{avgshade!75}\textbf{25.50} & \cellcolor{passshade!75}\textbf{41.89} & \cellcolor{lenshade!75}\textbf{1875} \\
\bottomrule
\end{tabular}
\end{adjustbox}
\end{subtable}

\vspace{0.5em}

\begin{subtable}{\linewidth}
\centering
\scriptsize
\setlength{\tabcolsep}{1.0pt}
\renewcommand{\arraystretch}{1.05}
\caption{Teacher: Qwen3-8B $\to$ Student: Qwen3-0.6B-Base}
\label{tab:main-8b}
\begin{adjustbox}{max width=\linewidth}
\begin{tabular}{@{}ll lll@{\hskip 6pt}lll@{\hskip 6pt}lll@{\hskip 6pt}lll@{\hskip 6pt}lll@{\hskip 6pt}lll@{\hskip 6pt}lll@{\hskip 6pt}lll@{}}
\toprule
 &  & \multicolumn{3}{c}{GSM8K} & \multicolumn{3}{c}{MATH500} & \multicolumn{3}{c}{AMC23} & \multicolumn{3}{c}{AIME24} & \multicolumn{3}{c}{AIME25} & \multicolumn{3}{c}{Minerva} & \multicolumn{3}{c}{Olympiad} & \multicolumn{3}{c}{Mean} \\
Class & Param & Avg & Pass & Len & Avg & Pass & Len & Avg & Pass & Len & Avg & Pass & Len & Avg & Pass & Len & Avg & Pass & Len & Avg & Pass & Len & Avg & Pass & Len \\
\cmidrule(lr){3-5} \cmidrule(lr){6-8} \cmidrule(lr){9-11} \cmidrule(lr){12-14} \cmidrule(lr){15-17} \cmidrule(lr){18-20} \cmidrule(lr){21-23} \cmidrule(lr){24-26}
\midrule
\multirow{1}{*}{Full-vocab} & -- & \textbf{70.85} & 88.48 & 4095 & \textbf{46.53} & 54.38 & 4095 & 27.50 & 35.00 & 4096 & \textbf{5.00} & 6.67 & 4096 & \textbf{3.33} & 6.67 & 4096 & \underline{11.58} & 15.07 & 4089 & 6.62 & 21.00 & 4096 & 24.49 & 32.47 & 4095 \\
\cmidrule(lr){1-26}
\multirow{4}{*}{Vanilla} & -- & 60.59 & 86.35 & 416 & 40.43 & 64.00 & 1450 & 22.19 & 50.00 & 2393 & 0.83 & 3.33 & 3436 & 0.83 & 6.67 & 3752 & 9.01 & 20.59 & 1306 & 14.35 & 33.33 & 2708 & 21.18 & 37.75 & 2209 \\
 & + Clip & 67.99 & 87.64 & 398 & 43.75 & 65.30 & 1429 & \underline{28.12} & 52.50 & 2538 & 1.67 & 6.67 & 3506 & 1.67 & \underline{10.00} & 3711 & 9.42 & 22.06 & 1286 & 16.20 & 34.22 & 2674 & 24.12 & 39.77 & 2220 \\
 & + Tanh & 66.98 & 88.63 & 394 & 43.39 & 64.86 & 1425 & 25.00 & \underline{57.50} & 2468 & 2.92 & \underline{10.00} & 3518 & 1.67 & 6.67 & 3639 & 9.65 & 21.32 & 1281 & 16.89 & 34.52 & 2698 & 23.79 & 40.50 & 2203 \\
 & + Z-Score & 63.83 & 88.10 & 391 & 42.72 & 65.62 & 1364 & 25.62 & 50.00 & 2269 & 1.25 & 6.67 & 3371 & 0.42 & 3.33 & 3578 & 10.71 & 21.32 & 1212 & 15.54 & 35.70 & 2528 & 22.87 & 38.68 & 2102 \\
\cmidrule(lr){1-26}
\multirow{8}{*}{PowerOPD} & $\alpha=0.1$ & 67.18 & 88.02 & 384 & 43.82 & 65.86 & 1395 & 24.06 & 55.00 & 2460 & \underline{3.75} & \textbf{13.33} & 3413 & 0.83 & 6.67 & 3617 & 10.25 & 22.43 & 1195 & 15.94 & 34.67 & 2632 & \cellcolor{avgshade!19}23.69 & \cellcolor{passshade!48}40.85 & \cellcolor{lenshade!21}2157 \\
 & $\alpha=0.5$ & 67.58 & 89.16 & 381 & 45.07 & 66.98 & 1371 & 25.94 & 55.00 & 2535 & 1.67 & 6.67 & 3359 & 0.83 & 6.67 & 3549 & 10.06 & 22.79 & 1144 & \textbf{17.26} & \textbf{36.89} & 2552 & \cellcolor{avgshade!34}24.06 & \cellcolor{passshade!35}40.59 & \cellcolor{lenshade!30}2127 \\
 & $\alpha=1$ & 65.36 & \underline{89.76} & \underline{380} & 44.89 & \textbf{67.10} & 1408 & 23.44 & 52.50 & 2560 & 1.67 & 6.67 & 3584 & 0.83 & 3.33 & 3542 & 11.31 & \textbf{25.37} & 1188 & \underline{17.24} & \underline{36.30} & 2637 & \cellcolor{avgshade!12}23.53 & \cellcolor{passshade!12}40.15 & \cellcolor{lenshade!12}2186 \\
 & $\alpha=5$ & 66.79 & 88.86 & \textbf{364} & 45.33 & 66.72 & 1348 & 25.62 & 55.00 & 2503 & 2.92 & 6.67 & 3512 & \textbf{3.33} & \textbf{13.33} & 3468 & 11.08 & 22.79 & \underline{1076} & 17.13 & 35.11 & 2566 & \cellcolor{avgshade!56}24.60 & \cellcolor{passshade!67}41.21 & \cellcolor{lenshade!39}2120 \\
 & $\alpha=10$ & 66.76 & 87.87 & \textbf{364} & \underline{45.49} & \underline{67.00} & \textbf{1308} & 24.38 & \textbf{60.00} & 2251 & 1.25 & 6.67 & 3274 & 0.83 & 6.67 & 3528 & \textbf{11.76} & \underline{24.63} & \textbf{1056} & \underline{17.24} & 36.00 & 2523 & \cellcolor{avgshade!30}23.96 & \cellcolor{passshade!69}41.26 & \cellcolor{lenshade!48}2043 \\
 & $\alpha=50$ & 69.29 & 88.93 & 382 & 44.91 & 65.78 & \underline{1313} & \underline{28.12} & \underline{57.50} & 2088 & 1.67 & 6.67 & 3262 & 0.42 & 6.67 & 3233 & 11.12 & 23.90 & 1149 & 16.96 & 35.70 & 2449 & \cellcolor{avgshade!58}24.64 & \cellcolor{passshade!42}40.74 & \cellcolor{lenshade!57}1982 \\
 & $\alpha=100$ & 68.63 & \textbf{89.79} & 385 & 45.03 & 65.42 & 1322 & \textbf{29.06} & \underline{57.50} & \underline{2038} & 2.92 & \underline{10.00} & \underline{3164} & 1.25 & \underline{10.00} & \underline{3225} & 11.31 & 22.79 & 1244 & 17.17 & 34.07 & \underline{2331} & \cellcolor{avgshade!75}\textbf{25.05} & \cellcolor{passshade!75}\textbf{41.37} & \cellcolor{lenshade!66}\underline{1958} \\
 & $\alpha=500$ & \underline{70.58} & 87.04 & 427 & 44.59 & 65.42 & 1321 & 22.19 & \underline{57.50} & \textbf{1952} & \textbf{5.00} & 6.67 & \textbf{3059} & \underline{2.08} & \textbf{13.33} & \textbf{2742} & 11.44 & 23.90 & 1309 & 16.83 & 35.26 & \textbf{2201} & \cellcolor{avgshade!59}\underline{24.67} & \cellcolor{passshade!71}\underline{41.30} & \cellcolor{lenshade!75}\textbf{1859} \\
\bottomrule
\end{tabular}
\end{adjustbox}
\end{subtable}

\vspace{-4pt}
\caption{
Mathematical reasoning evaluation with the Qwen3-0.6B student. \textbf{Bold} and \underline{underline} denote the best and second-best results. Color intensity highlights the relative performance of PowerOPD variants across different \(\alpha\). Log reward denotes OPD reward variants without post-hoc stabilization.
Avg, Pass, and Len denote Avg@8, Pass@8, and average response length, respectively.
}
\label{tab:main}
\end{table*}

The high variance of OPD rewards originates from the unboundedness of the log-ratio form. The OPD reward \(r_t^{\mathrm{OPD}}=\log(\pi_T/\pi_\theta)\) diverges in both directions\footnote{Full-vocabulary OPD shares the same unbounded log-ratio, but 
computes an exact expectation where extreme values are suppressed by 
their small probability weights. Sampled-token OPD lacks this 
averaging, exposing the gradient directly to extreme log-ratios.}: \(r_t^{\mathrm{OPD}}\to +\infty\) when \(\pi_\theta(o_t\mid c_t)\to 0\), and \(r_t^{\mathrm{OPD}}\to -\infty\) when \(\pi_T(o_t\mid c_t)\to 0\). Because OPD evaluates rewards on student-sampled tokens, these tokens tend to have high probability under the student, but the teacher may assign low probability to the same student-sampled tokens. The log-ratio reward can therefore easily enter its negative divergent regime, producing the heavy negative tail observed in \autoref{fig:reward-diagnosis}(a).
A standard remedy is to apply post-hoc reward transformations from RL, where clipping and normalization are widely used to stabilize optimization \citep{mnih2015human,andrychowicz2020matters}. We consider three representative methods, summarized in \autoref{tab:standard-fixes}: \emph{Clip} truncates rewards at fixed thresholds,\footnote{We set \(c_l=-c_h\), grid-search multiple threshold values, and use the best-performing setting \(c_l=-1, c_h=1\).} \emph{Tanh} maps rewards smoothly into \([-1,1]\), and \emph{Z-Score} centers and rescales rewards using batch statistics. 

\paragraph{Post-hoc stabilization fails to fix.}
As shown in \autoref{fig:standard-fixes}, post-hoc reward stabilization alleviates neither the optimization delay nor the unstable generation dynamics of OPD. \emph{Tanh} and \emph{Z-Score} reduce the gap to full-vocabulary OPD to some extent, but validation accuracy still improves only after a long delay, and response length continues to exhibit large oscillations. \emph{Z-Score} can even underperform vanilla OPD, likely because batch centering may change the sign of rewards and reverse the intended direction of some updates. In contrast, \emph{Clip} and \emph{Tanh} preserve the reward sign, but they only reshape reward magnitudes after the unbounded log-ratio has been computed, leaving the underlying reward form unchanged. These results indicate that \emph{OPD's pathological training dynamics are driven not by the lack of post-hoc stabilization, but by the unbounded log-ratio reward itself.}

\section{Rethinking the OPD Reward Function}
\label{sec:method}

As shown in \secref{sec:failure-modes}, post-hoc transformations of the log-ratio reward do not resolve OPD training pathologies. We therefore step back from the inherited log-ratio form and ask: \emph{what properties should a token-level OPD reward satisfy, and what functions satisfy them by construction?}

\subsection{Generalizing the OPD Reward}
\label{sec:framework}

The OPD log-ratio reward \(\log(\pi_T/\pi_\theta)\) is a specific mapping from teacher--student probabilities at the sampled token; we generalize it as
\begin{equation}
  r_t^{f} = f\!\left(\pi_T(o_t \mid c_t),\; \pi_\theta(o_t \mid c_t)\right),
\end{equation}
where \(f : [0,1]^2 \to \mathbb{R}\) maps a pair of teacher--student probabilities to a scalar reward. The standard OPD reward corresponds to \(f(p,q)=\log p-\log q\), which is defined on \((0,1]^2\) but diverges as \(p \to 0\) or \(q \to 0\). The post-hoc methods in \secref{subsec:standard-fixes} keep this log-ratio form fixed and only transform or normalize the rewards after they are computed. In contrast, \emph{we treat the probability-to-reward mapping \(f\) itself as the object of design}. This shifts the question from how to stabilize a pathological reward after the fact to what properties a stable OPD reward should satisfy in the first place.

\subsection{Two Necessary Properties}
\label{sec:properties}

We identify two necessary properties for a well-designed OPD reward.

\paragraph{P1: Boundedness.}
The reward function \(f\) should be bounded on \([0,1]^2\): there exists \(M > 0\) such that \(|f(p, q)| \leq M\) for all \(p, q \in [0,1]\). If \(f\) is unbounded, small changes in teacher or student probabilities can produce arbitrarily large reward magnitudes, directly amplifying the variance of policy-gradient updates as diagnosed in \secref{subsec:high-variance}.

\paragraph{P2: Sign Consistency.}
The sign of \(f\) should indicate which model assigns higher probability to the sampled token:
\(f(p,q)>0\) if \(p>q\), \(f(p,q)=0\) if \(p=q\), and \(f(p,q)<0\) if \(p<q\).
This ensures that token-level updates move the student toward the teacher: tokens assigned higher probability by the teacher receive positive reward, and vice versa.
If P2 is violated, updates can point in the wrong direction regardless of reward magnitude.

\paragraph{P1 and P2 explain the empirical failures.}
The failure modes in \secref{sec:failure-modes} are consistent with these two properties. \emph{Vanilla OPD} satisfies P2 but violates P1, producing the high-variance rewards diagnosed above. \emph{Z-Score} does not guarantee P1 and may violate P2 because batch centering can flip individual reward signs, explaining its poor performance in \autoref{fig:standard-fixes}. \emph{Clip} and \emph{Tanh} enforce P1 while preserving P2, and therefore improve over vanilla OPD; however, they still fail because they compress rewards only after the divergent log-ratio has been computed. Thus, P1 and P2 are minimal properties: they must hold at the probability-to-reward mapping level, before any log-ratio distortion occurs.
\subsection{Deriving Rewards that Satisfy P1 and P2}
\label{sec:motivation}
\begin{table*}[t]
\centering

\begin{subtable}{\linewidth}
\centering
\scriptsize
\setlength{\tabcolsep}{1.0pt}
\renewcommand{\arraystretch}{1.05}
\caption{Teacher: Qwen3-4B $\to$ Student: Qwen3-1.7B-Base}
\label{tab:main-1p7b-4b}
\begin{adjustbox}{max width=\linewidth}
\begin{tabular}{@{}ll lll@{\hskip 6pt}lll@{\hskip 6pt}lll@{\hskip 6pt}lll@{\hskip 6pt}lll@{\hskip 6pt}lll@{\hskip 6pt}lll@{\hskip 6pt}lll@{}}
\toprule
 &  & \multicolumn{3}{c}{GSM8K} & \multicolumn{3}{c}{MATH500} & \multicolumn{3}{c}{AMC23} & \multicolumn{3}{c}{AIME24} & \multicolumn{3}{c}{AIME25} & \multicolumn{3}{c}{Minerva} & \multicolumn{3}{c}{Olympiad} & \multicolumn{3}{c}{Mean} \\
Class & Param & Avg & Pass & Len & Avg & Pass & Len & Avg & Pass & Len & Avg & Pass & Len & Avg & Pass & Len & Avg & Pass & Len & Avg & Pass & Len & Avg & Pass & Len \\
\cmidrule(lr){3-5} \cmidrule(lr){6-8} \cmidrule(lr){9-11} \cmidrule(lr){12-14} \cmidrule(lr){15-17} \cmidrule(lr){18-20} \cmidrule(lr){21-23} \cmidrule(lr){24-26}
\midrule
\multirow{1}{*}{Full-vocab} & -- & \textbf{83.50} & 95.00 & 3954 & \underline{63.12} & \textbf{87.00} & 4004 & 38.44 & 65.00 & 4096 & \textbf{10.83} & \underline{23.33} & 4096 & 5.83 & \underline{20.00} & 4096 & \textbf{19.75} & \underline{32.00} & 3805 & 17.62 & 26.00 & 4096 & 34.16 & 49.76 & 4021 \\
\cmidrule(lr){1-26}
\multirow{4}{*}{Vanilla} & -- & 73.93 & \textbf{95.68} & 351 & 54.93 & 75.12 & 1156 & 31.87 & 60.00 & 2253 & 9.17 & 16.67 & 3388 & \textbf{7.08} & 13.33 & 3282 & 16.13 & 28.68 & 1038 & 23.74 & 44.15 & 2355 & 30.98 & 47.66 & 1975 \\
 & + Clip & 80.36 & 94.01 & 326 & 59.12 & 75.92 & 1162 & 31.88 & 67.50 & 2309 & 4.17 & \underline{23.33} & 3342 & 4.58 & \underline{20.00} & 3381 & 17.56 & 28.31 & 995 & 26.74 & \underline{47.41} & 2390 & 32.06 & 50.93 & 1986 \\
 & + Tanh & 82.08 & 94.62 & 331 & 59.09 & 76.04 & 1134 & 34.06 & 60.00 & 2288 & \underline{10.42} & \textbf{26.67} & 3318 & \underline{6.67} & 16.67 & 3240 & 18.06 & 30.88 & 926 & 27.17 & 46.96 & 2359 & 33.94 & 50.26 & 1942 \\
 & + Z-Score & 42.92 & 89.61 & 372 & 28.24 & 66.24 & 1235 & 15.94 & 47.50 & 2243 & 4.17 & 13.33 & 3466 & 4.17 & 13.33 & 3244 & 9.05 & 23.16 & 1062 & 14.15 & 38.81 & 2376 & 16.95 & 41.71 & 2000 \\
\cmidrule(lr){1-26}
\multirow{7}{*}{PowerOPD} & $\alpha=0.1$ & 82.46 & 94.77 & 316 & 59.66 & 76.56 & 1127 & 33.44 & 67.50 & 2184 & 8.75 & 20.00 & 3351 & \textbf{7.08} & \textbf{23.33} & 3081 & 18.11 & 29.41 & 912 & \textbf{27.76} & 47.26 & 2332 & \cellcolor{avgshade!12}33.89 & \cellcolor{passshade!49}\underline{51.26} & \cellcolor{lenshade!33}1900 \\
 & $\alpha=1$ & 82.04 & \underline{95.38} & 317 & 59.52 & \underline{76.86} & 1132 & 38.44 & 62.50 & 2176 & \underline{10.42} & \underline{23.33} & 3368 & 4.17 & 16.67 & 3136 & 18.66 & \textbf{32.35} & 935 & 27.72 & \textbf{47.56} & 2307 & \cellcolor{avgshade!36}34.42 & \cellcolor{passshade!30}50.66 & \cellcolor{lenshade!22}1910 \\
 & $\alpha=5$ & 82.68 & 95.07 & 315 & 59.96 & 76.42 & 1116 & 38.12 & 67.50 & 2136 & 8.33 & 20.00 & 3366 & 4.17 & 13.33 & 3283 & 19.07 & 31.25 & 920 & 27.28 & 47.11 & 2289 & \cellcolor{avgshade!28}34.23 & \cellcolor{passshade!12}50.10 & \cellcolor{lenshade!12}1918 \\
 & $\alpha=10$ & 83.01 & 94.69 & 315 & 59.71 & 76.10 & 1104 & 35.62 & 67.50 & 2112 & 8.33 & \textbf{26.67} & 3292 & 5.00 & 13.33 & 3188 & 18.84 & 30.15 & 909 & 27.56 & 46.67 & 2296 & \cellcolor{avgshade!18}34.01 & \cellcolor{passshade!32}50.73 & \cellcolor{lenshade!44}1888 \\
 & $\alpha=50$ & 83.11 & 94.24 & \underline{313} & 59.40 & 75.68 & 1035 & \textbf{41.88} & \textbf{75.00} & 1950 & \textbf{10.83} & \underline{23.33} & 3227 & 4.17 & 13.33 & 2876 & 17.65 & 30.15 & \underline{827} & 26.67 & 46.52 & 2160 & \cellcolor{avgshade!55}\underline{34.82} & \cellcolor{passshade!47}51.18 & \cellcolor{lenshade!54}1770 \\
 & $\alpha=100$ & 82.92 & 93.33 & \textbf{307} & 59.80 & 75.76 & \underline{1009} & 37.81 & \underline{70.00} & \underline{1857} & 8.33 & 20.00 & \underline{3011} & 6.25 & \textbf{23.33} & \underline{2732} & 17.97 & 28.31 & \textbf{823} & 27.06 & 45.78 & \underline{2042} & \cellcolor{avgshade!31}34.31 & \cellcolor{passshade!39}50.93 & \cellcolor{lenshade!64}\underline{1683} \\
 & $\alpha=500$ & \underline{83.13} & 94.31 & 316 & \textbf{64.53} & 75.72 & \textbf{997} & \underline{40.94} & \underline{70.00} & \textbf{1673} & 9.58 & \textbf{26.67} & \textbf{2776} & \underline{6.67} & \underline{20.00} & \textbf{2588} & \underline{19.58} & 31.25 & 845 & \underline{27.74} & 46.37 & \textbf{1857} & \cellcolor{avgshade!75}\textbf{36.02} & \cellcolor{passshade!75}\textbf{52.05} & \cellcolor{lenshade!75}\textbf{1579} \\
\bottomrule
\end{tabular}
\end{adjustbox}
\end{subtable}

\vspace{0.5em}

\begin{subtable}{\linewidth}
\centering
\scriptsize
\setlength{\tabcolsep}{1.0pt}
\renewcommand{\arraystretch}{1.05}
\caption{Teacher: Qwen3-8B $\to$ Student: Qwen3-1.7B-Base}
\label{tab:main-1p7b-8b}
\begin{adjustbox}{max width=\linewidth}
\begin{tabular}{@{}ll lll@{\hskip 6pt}lll@{\hskip 6pt}lll@{\hskip 6pt}lll@{\hskip 6pt}lll@{\hskip 6pt}lll@{\hskip 6pt}lll@{\hskip 6pt}lll@{}}
\toprule
 &  & \multicolumn{3}{c}{GSM8K} & \multicolumn{3}{c}{MATH500} & \multicolumn{3}{c}{AMC23} & \multicolumn{3}{c}{AIME24} & \multicolumn{3}{c}{AIME25} & \multicolumn{3}{c}{Minerva} & \multicolumn{3}{c}{Olympiad} & \multicolumn{3}{c}{Mean} \\
Class & Param & Avg & Pass & Len & Avg & Pass & Len & Avg & Pass & Len & Avg & Pass & Len & Avg & Pass & Len & Avg & Pass & Len & Avg & Pass & Len & Avg & Pass & Len \\
\cmidrule(lr){3-5} \cmidrule(lr){6-8} \cmidrule(lr){9-11} \cmidrule(lr){12-14} \cmidrule(lr){15-17} \cmidrule(lr){18-20} \cmidrule(lr){21-23} \cmidrule(lr){24-26}
\midrule
\multirow{1}{*}{Full-vocab} & -- & 80.25 & 86.50 & 4095 & \textbf{63.25} & \textbf{84.00} & 4091 & \textbf{42.19} & \underline{75.00} & 4096 & 7.08 & 16.67 & 4096 & \textbf{5.83} & \textbf{26.67} & 4096 & 17.50 & 26.00 & 4084 & 16.50 & 29.00 & 4096 & 33.23 & 49.12 & 4093 \\
\cmidrule(lr){1-26}
\multirow{4}{*}{Vanilla} & -- & 77.55 & \underline{95.07} & 323 & 57.55 & 75.88 & 1050 & 35.94 & 62.50 & 1881 & 7.08 & 20.00 & 3045 & 3.30 & 16.67 & 2957 & 17.56 & 29.78 & 892 & 23.94 & 47.11 & 2136 & 31.85 & 49.57 & 1755 \\
 & + Clip & 76.96 & 94.84 & \textbf{308} & 59.56 & 76.68 & 1062 & 38.75 & 72.50 & 2097 & 7.08 & \underline{23.33} & 3188 & 3.30 & 16.67 & 2935 & 18.43 & 30.88 & 892 & 24.65 & 46.81 & 2177 & 32.68 & 51.67 & 1808 \\
 & + Tanh & 78.02 & 94.92 & 321 & 58.82 & 76.04 & 1078 & 35.31 & 62.50 & 1996 & \underline{8.33} & \underline{23.33} & 3107 & 4.58 & \underline{20.00} & 3012 & 18.29 & \underline{31.62} & 930 & 23.96 & 47.41 & 2205 & 32.47 & 50.83 & 1807 \\
 & + Z-Score & 80.26 & \textbf{95.22} & 323 & 58.48 & 75.98 & 1084 & 36.56 & 67.50 & 2087 & 7.08 & \underline{23.33} & 3285 & 4.17 & 13.33 & 2964 & 17.42 & \textbf{31.99} & 948 & 27.02 & \underline{48.00} & 2174 & 33.00 & 50.76 & 1838 \\
\cmidrule(lr){1-26}
\multirow{7}{*}{PowerOPD} & $\alpha=0.1$ & 79.96 & 94.31 & 319 & 59.27 & 76.78 & 1087 & 36.25 & 65.00 & 2142 & \textbf{9.17} & \underline{23.33} & 3123 & 3.75 & 13.33 & 3024 & 18.43 & \textbf{31.99} & 920 & 27.09 & 47.26 & 2227 & \cellcolor{avgshade!33}33.42 & \cellcolor{passshade!12}50.29 & \cellcolor{lenshade!12}1835 \\
 & $\alpha=0.5$ & 78.22 & 94.77 & \textbf{308} & 59.29 & 76.90 & 1052 & 36.56 & 67.50 & 2087 & \underline{8.33} & \textbf{26.67} & 3195 & \underline{5.42} & \underline{20.00} & 2962 & 17.97 & 30.88 & 899 & \underline{27.78} & \underline{48.00} & 2143 & \cellcolor{avgshade!31}33.37 & \cellcolor{passshade!75}\textbf{52.10} & \cellcolor{lenshade!25}1807 \\
 & $\alpha=1$ & 76.21 & 94.31 & \underline{311} & 59.04 & 76.90 & 1083 & 37.50 & 70.00 & 2113 & \underline{8.33} & 20.00 & 3200 & 5.00 & 16.67 & 3019 & 17.46 & 29.78 & 924 & 26.81 & 46.52 & 2192 & \cellcolor{avgshade!12}32.91 & \cellcolor{passshade!23}50.60 & \cellcolor{lenshade!12}1835 \\
 & $\alpha=5$ & 79.54 & 94.16 & 312 & 59.82 & \underline{77.12} & 1062 & \underline{40.31} & 70.00 & 2010 & 7.08 & 20.00 & 3198 & 3.75 & 13.33 & 2939 & 18.75 & 30.88 & 880 & 27.48 & \textbf{48.15} & 2202 & \cellcolor{avgshade!49}33.82 & \cellcolor{passshade!20}50.52 & \cellcolor{lenshade!37}1800 \\
 & $\alpha=10$ & 78.79 & 93.93 & 321 & 59.86 & 76.76 & 1049 & 38.75 & 62.50 & 1997 & \underline{8.33} & \textbf{26.67} & 3206 & 4.58 & 16.67 & \underline{2853} & 18.89 & \underline{31.62} & 882 & \textbf{27.81} & 46.07 & 2148 & \cellcolor{avgshade!51}33.86 & \cellcolor{passshade!23}50.60 & \cellcolor{lenshade!50}1779 \\
 & $\alpha=100$ & \underline{83.06} & 93.93 & 319 & 59.36 & 75.74 & \textbf{998} & 38.75 & \textbf{77.50} & \underline{1821} & \textbf{9.17} & \underline{23.33} & \underline{2869} & 3.33 & 13.33 & 2928 & \textbf{19.07} & 29.41 & \textbf{826} & 26.72 & 45.48 & \underline{2008} & \cellcolor{avgshade!65}\underline{34.21} & \cellcolor{passshade!45}51.25 & \cellcolor{lenshade!62}\underline{1681} \\
 & $\alpha=500$ & \textbf{83.85} & 94.16 & 322 & \underline{61.06} & 75.90 & \underline{999} & 36.56 & \textbf{77.50} & \textbf{1715} & \underline{8.33} & \underline{23.33} & \textbf{2705} & \textbf{5.83} & 16.67 & \textbf{2557} & \underline{18.93} & 29.78 & \underline{846} & 26.56 & 45.78 & \textbf{1827} & \cellcolor{avgshade!75}\textbf{34.45} & \cellcolor{passshade!67}\underline{51.87} & \cellcolor{lenshade!75}\textbf{1567} \\
\bottomrule
\end{tabular}
\end{adjustbox}
\end{subtable}

\vspace{-4pt}
\caption{
Mathematical reasoning evaluation with the Qwen3-1.7B student. \textbf{Bold} and \underline{underline} denote the best and second-best results. Color intensity highlights the relative performance of PowerOPD variants across different \(\alpha\). Vanilla denotes OPD reward variants without post-hoc stabilization.
Avg, Pass, and Len denote Avg@8, Pass@8, and average response length, respectively.}
\label{tab:main-1p7b}
\end{table*}

\begin{figure*}[t]
  \centering
  \begin{minipage}[t]{0.245\textwidth}
    \centering
    \includegraphics[width=\linewidth]{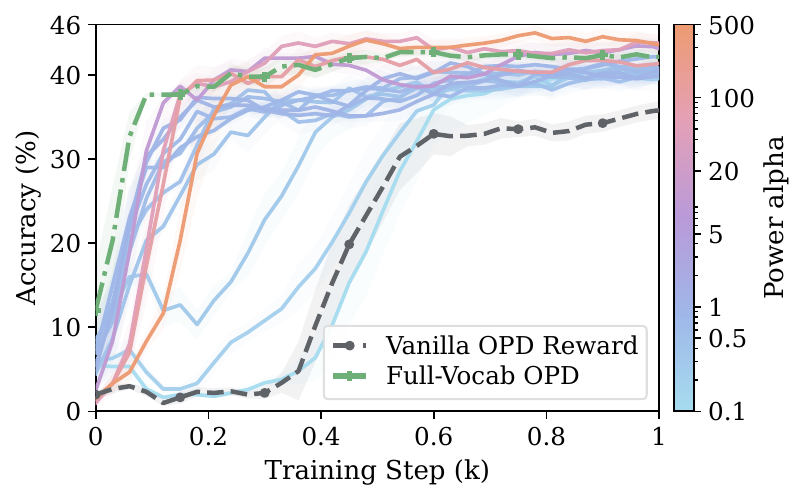}\\[-0.5ex]
    \textbf{(a)}
  \end{minipage}\hfill
  \begin{minipage}[t]{0.245\textwidth}
    \centering
    \includegraphics[width=\linewidth]{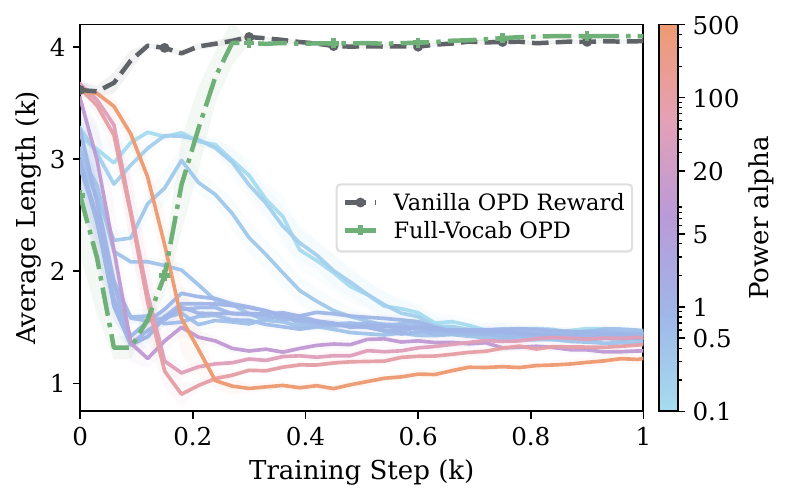}\\[-0.5ex]
    \textbf{(b)}
  \end{minipage}\hfill
  \begin{minipage}[t]{0.245\textwidth}
    \centering
    \includegraphics[width=\linewidth]{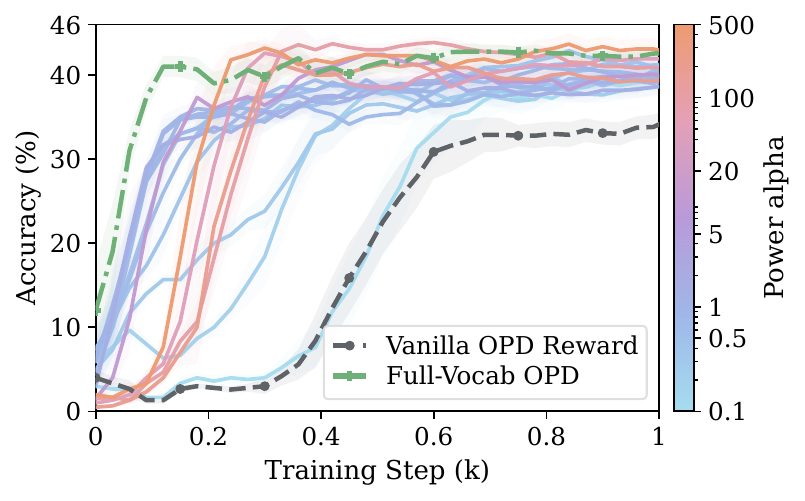}\\[-0.5ex]
    \textbf{(c)}
  \end{minipage}\hfill
  \begin{minipage}[t]{0.245\textwidth}
    \centering
    \includegraphics[width=\linewidth]{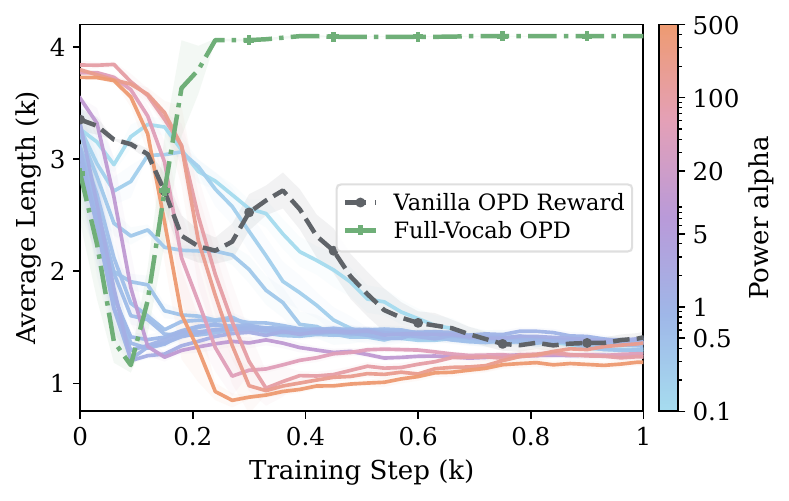}\\[-0.5ex]
    \textbf{(d)}
  \end{minipage}
  \vspace{-7pt}
    \caption{
    Training dynamics across \(\alpha\) for a Qwen3-0.6B-Base student: (a,b) accuracy and length with a Qwen3-4B teacher; (c,d) accuracy and length with a Qwen3-8B teacher.
    }
    \vspace{-10pt}
  \label{fig:alpha}
\end{figure*}

We now identify what structure a natively bounded and sign-consistent reward should take.
To obtain a tractable family, we observe that the log-ratio reward has a specific algebraic structure: it applies the same scalar transformation to each probability, then takes their difference, giving $\log\pi_T - \log\pi_\theta = h(\pi_T) - h(\pi_\theta)$ where $h(x) = \log x$.
This \emph{transform-then-subtract} class \(f_h(p,q) = h(p) - h(q)\) is attractive because sign consistency follows immediately whenever \(h\) is strictly monotone increasing: if \(p > q\), then \(h(p) > h(q)\) and \(r_t > 0\). Thus, P2 is guaranteed by construction.
The stability of the reward is then controlled entirely by the choice of $h$: the problem with the log ratio is not the transform-then-subtract structure, but the specific choice $h(x) = \log x$, which maps $[0,1]$ to $(-\infty, 0]$ and diverges at zero, making $h(\pi_T) - h(\pi_\theta)$ unbounded and violating P1.
The fix is therefore precise: keep the transform-then-subtract structure and replace $h = \log$ with a function that is both strictly monotone increasing and bounded on $[0,1]$.

\paragraph{The Box--Cox family.}
A natural and well-studied family for this purpose is the Box--Cox power transformation~\citep{boxcox1964},
\begin{equation}
  h_\alpha(x) = \frac{x^\alpha - 1}{\alpha}, \quad \alpha > 0.
  \label{eq:boxcox}
\end{equation}
For any \(\alpha>0\), \(h_\alpha\) is strictly increasing and bounded on \([0,1]\). Substituting \(h_\alpha\) into \(h(p)-h(q)\) gives \((p^\alpha-q^\alpha)/\alpha\). Since \(1/\alpha\) is a positive constant independent of tokens and policy parameters, we absorb it into the learning rate and use the rescaled reward \(p^\alpha-q^\alpha\), which is bounded in \([-1,1]\).

\paragraph{The log ratio as a limiting case.}
The standard log-ratio reward corresponds to the boundary case \(\alpha \to 0\) of the Box--Cox transformation. By the first-order Taylor expansion \(x^\alpha = 1+\alpha\log x+o(\alpha)\), we have \(h_\alpha(x)=(x^\alpha-1)/\alpha \to \log x\), and therefore \(h_\alpha(p)-h_\alpha(q)\to \log p-\log q\). However, this limit is degenerate for reward design: although each fixed \(\alpha>0\) gives a bounded transformation, its range expands as \(\alpha\to0\), recovering the unbounded log transformation and thereby violating P1.

\subsection{PowerOPD}
\label{sec:poweropd}

The analysis above identifies a principled family of natively bounded, sign-consistent OPD rewards. Following the policy-gradient formulation in \secref{sec:opd-as-rl}, we introduce \textbf{PowerOPD}, which uses the following stop-gradient token-level reward:
\begin{equation}
  r_t^{\alpha} = \mathrm{sg}\!\left[\pi_T(o_t \mid c_t)^{\alpha} - \pi_\theta(o_t \mid c_t)^{\alpha}\right], \quad \alpha > 0.
  \label{eq:poweropd}
\end{equation}

\paragraph{Verification.}
Since \(p,q\in[0,1]\) and \(\alpha>0\), we have \(p^\alpha,q^\alpha\in[0,1]\), so \(r_t^\alpha\in[-1,1]\): \textbf{P1 holds}.
Since \(h(x)=x^\alpha\) is strictly increasing for \(\alpha>0\), \(\pi_T>\pi_\theta \Rightarrow r_t^\alpha>0\): \textbf{P2 holds}.
Critically, both hold at the probability-to-reward mapping level, without passing through the log ratio.

\paragraph{\(\alpha\) shapes the reward sensitivity profile.}
The exponent \(\alpha\) controls how reward sensitivity is distributed across the probability range. \emph{Smaller \(\alpha\) gives relatively more sensitivity to low-probability tokens}, resembling the log ratio's emphasis on rare events while remaining bounded. \emph{Larger \(\alpha\) attenuates low-probability differences} and shifts the reward signal toward better-supported regions.
\section{Experimental Setup}
\label{subsec:experimental-setup}

\paragraph{Models.}
We evaluate four teacher--student pairs spanning two student sizes and two teacher sizes: Qwen3-0.6B-Base and Qwen3-1.7B-Base as students, and Qwen3-4B and Qwen3-8B as teachers.

\paragraph{Training.}
We train on DeepScaleR~\citep{deepscaler2025} for 1.5k steps with bf16, learning rate \(5\times10^{-7}\), batch size 32, and on-policy rollouts up to 1024 tokens. Results are averaged over three runs.

\paragraph{Evaluation.}
We evaluate on six mathematical reasoning benchmarks: AIME24/25~\citep{maa_aime}, MATH-500~\citep{hendrycks2021math}, AMC23~\citep{maa_amc}, Minerva~\citep{minerva}, and OlympiadBench~\citep{OlympiadBench}. We report Avg@\(n\), Pass@\(n\), and average response length with \(n=8\).

\paragraph{Baselines.}
Baselines include \emph{Vanilla OPD}, reward-stabilized variants (\emph{Clip}: \(c_l=-1,c_h=1\); \emph{Tanh}; \textbf{Z-Score}), and \emph{Full-vocabulary KL OPD}.
\section{Experiments}
\label{sec:experiments}

\subsection{Main Results}
\label{sec:main-results}
 
\paragraph{PowerOPD substantially outperforms vanilla OPD.}
PowerOPD consistently improves over vanilla OPD across all four teacher--student configurations (Tables~\ref{tab:main} and~\ref{tab:main-1p7b}). Averaged over configurations using the best PowerOPD variant for each metric, PowerOPD improves the mean performance by \(\mathbf{+4.47}\) Avg@8 and \(\mathbf{+4.06}\) Pass@8. The largest benchmark-averaged gain occurs in the Qwen3-4B \(\to\) Qwen3-0.6B-Base setting, with Avg@8 improving from 19.13 to 25.50 (\(\mathbf{+6.37}\)) and Pass@8 from 36.18 to 41.89 (\(\mathbf{+5.71}\)). Gains also hold for the Qwen3-8B \(\to\) Qwen3-0.6B-Base setting (\(+3.87\) Avg@8, \(+3.62\) Pass@8), the Qwen3-4B \(\to\) Qwen3-1.7B-Base setting (\(+5.04\) Avg@8, \(+4.39\) Pass@8), and the Qwen3-8B \(\to\) Qwen3-1.7B-Base setting (\(+2.60\) Avg@8, \(+2.53\) Pass@8). On individual benchmarks, the gains are larger, reaching up to \(\mathbf{+16.75}\) Avg@8 on AMC23 in the Qwen3-4B \(\to\) Qwen3-0.6B-Base setting and \(\mathbf{+15.00}\) Pass@8 on AMC23 in the Qwen3-4B \(\to\) Qwen3-1.7B-Base and Qwen3-8B \(\to\) Qwen3-1.7B-Base settings. Notably, \emph{PowerOPD plugs directly into the standard OPD pipeline without modifying rollout generation, teacher scoring, or policy-gradient optimization}.

\begin{table}[t]
  \centering
  \footnotesize
  \setlength{\tabcolsep}{3pt}
  \begin{adjustbox}{max width=\columnwidth}
    \begin{tabular}{lccc}
    \toprule
    \textbf{Method} & \textbf{TFLOPs/upd.} & \textbf{Time/step} & \textbf{Peak mem.} \\
    \midrule
    Full-vocab OPD & 402.7 & 22.14s & 78.99 GiB \\
    PowerOPD & 346.6 & 9.03s & 60.72 GiB \\
    \midrule
    Reduction & $\downarrow$13.9\% & $\downarrow$59.2\% & $\downarrow$23.1\% \\
    \bottomrule
  \end{tabular}
  \end{adjustbox}
  \vspace{-5pt}
  \caption{Efficiency comparison on Qwen3-0.6B-Base $\leftarrow$ Qwen3-4B. Wall-time is averaged over 1.5k training steps; peak memory is measured with batch size 8.}
  \vspace{-12pt}
  \label{tab:efficiency}
\end{table}

\paragraph{PowerOPD also surpasses post-hoc stabilization methods.}
PowerOPD also outperforms post-hoc log-reward stabilization baselines (Tables~\ref{tab:main} and~\ref{tab:main-1p7b}). The largest benchmark-averaged gain occurs in the Qwen3-4B \(\to\) Qwen3-0.6B-Base setting, where PowerOPD improves over the strongest post-hoc baseline from 22.49 to 25.50 Avg@8 (\(\mathbf{+3.01}\)) and from 38.35 to 41.89 Pass@8 (\(\mathbf{+3.54}\)). At the individual-benchmark level, the gains reach up to \(\mathbf{+8.43}\) Avg@8 and \(\mathbf{+7.50}\) Pass@8 on AMC23 in the same setting. Notably, \emph{Z-Score} can underperform vanilla OPD, consistent with our analysis in \secref{sec:properties} that batch centering may violate sign consistency. These results show that post-hoc stabilization is insufficient: \emph{the reward properties must hold at the probability-to-reward mapping level, before the log-ratio distortion occurs.}

\begin{figure*}[t]
  \centering
  \includegraphics[width=\textwidth]{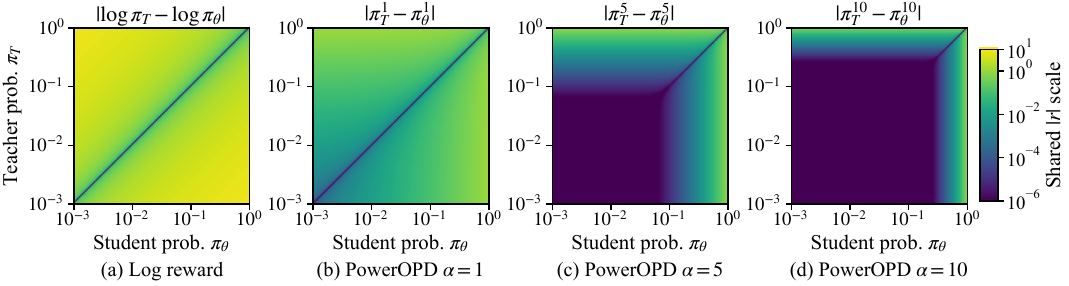}
    \caption{Reward magnitude over the joint probability space $(\pi_T, \pi_\theta)$. (a) The log-ratio reward depends only on the ratio $\pi_T/\pi_\theta$: it retains full sensitivity where both probabilities are small and grows without bound as either approaches zero. (b--d) PowerOPD couples reward magnitude to the absolute probability level: as $\alpha$ grows, the inert region expands and the signal contracts toward tokens with substantial probability under at least one model.}
  \label{fig:alpha-heatmap}
  \vspace{-10pt}
\end{figure*}

\begin{figure}[t]
  \centering
  \includegraphics[width=\columnwidth]{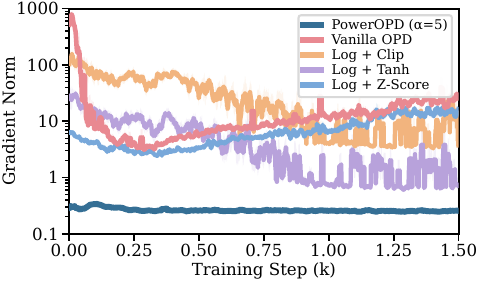}
    \caption{
    Gradient norm during training with a Qwen3-1.7B-Base student and Qwen3-4B teacher.
    }
    \label{fig:stability}
  
\end{figure}

    

\paragraph{PowerOPD consistently outperforms full-vocabulary OPD at substantially lower compute cost.}
Taking the best PowerOPD variant separately for each metric, the largest benchmark-averaged gains are \(\mathbf{+2.59}\) Avg@8 and \(\mathbf{+8.90}\) Pass@8. Specifically, with the Qwen3-0.6B-Base student, PowerOPD improves Avg@8 by \(+2.59\) and \(+0.56\) over full-vocabulary OPD under the 4B and 8B teachers, with corresponding Pass@8 gains of \(+8.17\) and \(+8.90\). With the Qwen3-1.7B-Base student, PowerOPD improves Avg@8 by \(+1.86\) and \(+1.22\), and Pass@8 by \(+2.29\) and \(+2.98\), under the 4B and 8B teachers, respectively. At the individual-benchmark level, the gains are substantially larger, reaching up to \(\mathbf{+11.60}\) Avg@8 on Olympiad in the Qwen3-4B \(\to\) Qwen3-0.6B-Base setting and \(\mathbf{+25.00}\) Pass@8 on AMC23 in the Qwen3-8B \(\to\) Qwen3-0.6B-Base setting. As shown in \autoref{tab:efficiency}, PowerOPD reduces wall-clock time by \(59.2\%\) and peak GPU memory by \(23.1\%\), with FLOPs computation detailed in \autoref{app:flops}. These results show that a bounded sampled-token reward can surpass full-vocabulary distillation while avoiding vocabulary-wide computation.
 
\paragraph{PowerOPD scales with larger \(\alpha\).}
Across all configurations, increasing \(\alpha\) generally improves Avg@8 and Pass@8 while shortening responses. In the 0.6B/4B setting, Avg@8 rises from 24.16 at \(\alpha{=}0.1\) to 25.50 at \(\alpha{=}500\), Pass@8 from 39.61 to 41.89, and average length drops from 2{,}291 to 1{,}875 tokens. The same trend holds in the 1.7B/4B setting, where Avg@8 improves from 33.89 to 36.02 and length decreases from 1{,}900 to 1{,}579 tokens. This suggests that larger \(\alpha\) encourages more targeted responses rather than longer generations. In contrast, full-vocabulary OPD often saturates the 4,096-token limit, consistent with the length inflation observed by \citet{luo2026demystifying}, 
indicating weaker length calibration despite strong accuracy. Notably, PowerOPD scales with \(\alpha\) without any additional supervision.

\subsection{Analysis}
\paragraph{Training Dynamics Across \texorpdfstring{\(\alpha\)}{alpha}}
\label{subsec:alpha-training-dynamics}

We conduct a fine-grained sweep over \(\alpha\) to study how the PowerOPD reward shape affects training dynamics. As shown in \autoref{fig:alpha}, we compare PowerOPD with vanilla OPD and full-vocabulary OPD under the Qwen3-0.6B-Base student with Qwen3-4B and Qwen3-8B teachers. First, \emph{larger \(\alpha\) leads to faster and stronger convergence.} Vanilla OPD exhibits the delayed-improvement phase diagnosed earlier and converges to a weaker final accuracy, whereas PowerOPD reaches high validation accuracy much earlier in training; larger \(\alpha\) further improves both convergence speed and final accuracy. Second, \emph{larger \(\alpha\) yields shorter and more stable generations.} As \(\alpha\) increases, PowerOPD drives the student toward shorter responses and stabilizes earlier, while vanilla OPD remains length-unstable and can fail to settle into a stable generation regime. Full-vocabulary OPD also often approaches the maximum generation length, indicating that stronger accuracy does not necessarily translate into better length calibration.

\paragraph{Why Larger \texorpdfstring{\(\alpha\)}{alpha} Improves PowerOPD}
\label{subsec:reward-sensitivity}

To understand why larger $\alpha$ improves accuracy, we examine how the reward distributes its magnitude over the joint probability space $(\pi_T, \pi_\theta)$ in \autoref{fig:alpha-heatmap}. The log-ratio reward depends only on the ratio $\pi_T/\pi_\theta$ and is blind to the absolute probability level: it assigns the same reward to $0.002$ versus $0.0001$ as to $0.2$ versus $0.01$, and grows without bound whenever one probability approaches zero while the other does not (\autoref{fig:alpha-heatmap}(a)). Consequently, its most extreme values fall on tokens that are improbable under both models, where the student rarely samples the token and the probability estimates carry the least statistical support, creating high-leverage policy-gradient terms $r_t \nabla_\theta \log \pi_\theta(o_t \mid c_t)$ from the least reliable measurements~\citep{williams1992reinforce, greensmith2004variance}. PowerOPD couples the reward magnitude to the absolute probability level: $|r_t^{\alpha}|$ is governed by $\max(\pi_T, \pi_\theta)^{\alpha}$, so a token receives substantial reward only if at least one model assigns it substantial probability, and $\alpha$ sets how strict this support requirement is. At $\alpha{=}1$, a probability of $0.6$ retains a sizable reward, whereas at $\alpha{=}10$ it contributes only $0.6^{10} \approx 0.006$. Accordingly, in \autoref{fig:alpha-heatmap}(b--d), the inert region expands as $\alpha$ grows, and the surviving signal contracts toward tokens where the teacher probability is high, so the distillation target is reliable, or the student probability is high, so the update adjusts the dominant modes of the current policy rather than its marginal behaviors. Larger $\alpha$ therefore suppresses exactly the unreliable high-leverage tokens that destabilize vanilla OPD (\secref{subsec:high-variance}), while concentrating learning on well-supported ones.

\paragraph{Training Stability}
\label{subsec:training-stability}

We further assess training stability by tracking gradient norms with a Qwen3-1.7B-Base student and Qwen3-4B teacher. As shown in \autoref{fig:stability}, vanilla OPD has an initial spike near \(10^3\) and later rises above \(20\), reflecting instability from high-variance log-ratio rewards. Post-hoc methods only partially mitigate this behavior: \emph{Clip} and \emph{Tanh} still fluctuate sharply, often exceeding \(10\), while \emph{Z-Score} grows from around \(3\) to above \(10\). In contrast, PowerOPD keeps the gradient norm nearly flat at \(0.25\)--\(0.35\), about \(3{,}000\times\) smaller than the initial OPD spike, over \(60\times\) smaller than late-stage OPD, and over \(30\times\) smaller than the high-gradient regimes of \emph{Clip} and \emph{Z-Score}. This shows that bounding the reward at the probability-to-reward mapping level stabilizes policy-gradient updates more effectively than post-hoc transformations of the log-ratio reward.
\section{Conclusion}
\label{sec:conclusion}
 
We show that OPD's training instability stems from the unbounded log-ratio reward, and that standard post-hoc fixes are insufficient.
We propose PowerOPD, a family of bounded, sign-consistent rewards from the Box-Cox power transformation parameterized by $\alpha > 0$, which consistently outperforms vanilla OPD and all post-hoc baselines, surpasses full-vocabulary OPD at substantially lower compute cost, and scales with $\alpha$ to simultaneously improve accuracy, shorten response length, and stabilize training dynamics throughout optimization.
\section*{Limitations}

Our experiments are conducted primarily on mathematical reasoning benchmarks and Qwen3-based teacher–student pairs. This setting provides a controlled testbed for studying OPD reward design, since it involves long-form generation and makes training instabilities easy to observe. Future work can further validate whether the same reward design brings similar benefits to tasks such as code generation, general instruction following, and multilingual reasoning.

\section*{Ethical Considerations}
This work studies the reward design of on-policy distillation for language model post-training. We do not introduce new datasets containing personal or sensitive information, nor do we propose a user-facing application or deployment pipeline. Therefore, we do not identify direct ethical concerns specific to the proposed method. More broadly, PowerOPD helps reduce the computational cost of distillation by improving the effectiveness of sampled-token training, which could make post-training more accessible and resource-efficient.


\bibliography{custom}
\newpage
\appendix

\section{Related Work}

\paragraph{On-Policy Distillation.}
Knowledge distillation  transfers knowledge from a teacher to a student by matching soft output distributions~\cite{hinton2015distilling,hsieh2023distilling,chen2025unveiling}.
For autoregressive language models, on-policy distillation trains the student on its own samples rather than on teacher-generated data, eliminating the train--test distribution mismatch of offline methods \citep{gu2024minillm, agarwal2024gkd}.
OPD has since been adopted at scale: Qwen3 \citep{qwen3} and GLM-5 \citep{glm5} use it to transfer reasoning capabilities, Nemotron-Cascade~2 \citep{nemotron2} combines it with cascade RL, and DeepSeek-V4 \citep{deepseekv4} integrates it into post-training.
\citet{song2026survey} survey the rapidly expanding landscape.

\paragraph{Training Stability in OPD.}
A growing body of work addresses OPD instability while retaining the log-ratio reward.
At the reward level, \citet{jin2026entropy} downweight uncertain tokens by teacher entropy, \citet{xu2026tip} assign importance weights based on reward reliability, and \citet{oh2026kl} introduce control variate baselines to reduce reward variance.
At the structural level, \citet{jang2026stable} adaptively reformulate the teacher target, \citet{jia2026asymmetric} apply asymmetric treatment across token positions, and \citet{hou2026uniopd} unify multiple distillation perspectives with stabilizing constraints.
Empirical analyses by \citet{fu2026revisiting}, \citet{luo2026demystifying}, and \citet{li2026rethinking} characterize common failure modes such as reward spikes and length inflation.
\citet{ko2026reopold} formalize OPD as a policy gradient problem and propose relaxed objectives.
All of these approaches modify how the log-ratio reward is weighted, normalized, or combined, but none question whether the log ratio is the right reward function.
Our work departs from this line by redesigning the reward itself, replacing the unbounded log transformation with bounded power functions derived from the Box-Cox family \citep{boxcox1964}.

\section{FLOPs Estimation}
\label{app:flops}

We estimate the distillation-update FLOPs for the Qwen3-0.6B-Base student and Qwen3-4B teacher setting. We use batch size \(B=32\), average prompt length \(128\), rollout length \(T=1024\), and total prefill length \(S=1152\). Let \(d\) be the hidden size, \(L\) the number of transformer layers, \(m\) the intermediate size, \(V\) the vocabulary size, and \(d_{\mathrm{kv}}=n_{\mathrm{kv}}d_h\) the total key/value projection dimension under GQA. For one prefill forward pass, we estimate the transformer FLOPs as
\[
\begin{aligned}
F_{\mathrm{tr}}(d,L,m,d_{\mathrm{kv}},S)
= BL \Big(
&4Sd^2 + 4Sdd_{\mathrm{kv}} \\
&+ 2S^2d + 6Sdm
\Big).
\end{aligned}
\]
The four terms correspond to the query/output projections, key/value projections, causal attention matrix multiplications, and SwiGLU MLP projections, respectively. We count multiply-add as two FLOPs. For Qwen3-4B, this gives
\[
F_{\mathrm{tr}}^{T} \approx 254.8\ \mathrm{TFLOPs},
\]
and for Qwen3-0.6B-Base,
\[
F_{\mathrm{tr}}^{S} \approx 30.6\ \mathrm{TFLOPs}.
\]
The teacher is used only for scoring, so we count one forward pass. The student is updated by backpropagation, so we approximate the student update as one forward pass plus backward pass, i.e., \(3F_{\mathrm{tr}}^{S}\). Thus, vanilla OPD methods such as PowerOPD require
\[
\begin{aligned}
F_{\mathrm{sampled}}
&= F_{\mathrm{tr}}^{T} + 3F_{\mathrm{tr}}^{S} \\
&\approx 254.8 + 3\times 30.6 \\
&= 346.6\ \mathrm{TFLOPs}.
\end{aligned}
\]
per distillation update. The sampled-token reward only requires the probabilities of the sampled rollout tokens; the corresponding selective output projection is \(O(BTd)\) and is negligible compared with transformer compute.

Full-vocabulary KL OPD additionally materializes vocabulary-sized distributions over all rollout positions. The teacher requires a full-vocabulary LM-head projection,
\[
F_{\mathrm{head}}^{T}
=
2BTd_TV
\approx
25.5\ \mathrm{TFLOPs},
\]
while the student full-vocabulary projection participates in backpropagation:
\[
F_{\mathrm{head}}^{S,\mathrm{train}}
=
3\cdot 2BTd_SV
\approx
30.6\ \mathrm{TFLOPs}.
\]
The KL term is computed over the full vocabulary,
\[
\begin{aligned}
D_{\mathrm{KL}}(\pi_\theta \| \pi_T)
=
\sum_{v=1}^{V}
\pi_\theta(v\mid c_t)
\Big[
&\log \pi_\theta(v\mid c_t) \\
&-\log \pi_T(v\mid c_t)
\Big].
\end{aligned}
\]
for each rollout token. This elementwise KL computation costs \(O(BTV)\), which is lower order relative to the full-vocabulary projections and is omitted from the headline estimate. Therefore, full-vocabulary KL OPD requires
\[
\begin{aligned}
F_{\mathrm{full}}
&= F_{\mathrm{tr}}^{T}
+ 3F_{\mathrm{tr}}^{S}
+ F_{\mathrm{head}}^{T}
+ F_{\mathrm{head}}^{S,\mathrm{train}} \\
&\approx 346.6 + 25.5 + 30.6 \\
&= 402.7\ \mathrm{TFLOPs}.
\end{aligned}
\]
Thus, avoiding the full-vocabulary distillation signal saves approximately
\[
402.7 - 346.6 = 56.1\ \mathrm{TFLOPs}
\]
per update, corresponding to a \(13.9\%\) reduction in distillation-update FLOPs. This estimate only counts arithmetic operations; in practice, full-vocabulary KL OPD also incurs additional memory traffic from materializing vocabulary-sized logits and KL tensors.

\section{Derivation of the OPD Policy-Gradient Form}
\label{app:opd-pg-derivation}

We derive the policy-gradient form used for OPD and clarify why the log-ratio term can be viewed as a dense token-level reward. Let \(x\sim\mathcal{D}\) be a prompt and \(o=(o_1,\ldots,o_T)\sim\pi_\theta(\cdot\mid x)\) be a student rollout. We denote the prefix context before token \(o_t\) by
\[
c_t = (x,o_{<t}).
\]
By autoregressive factorization,
\[
\begin{aligned}
\pi_\theta(o\mid x)
&= \prod_{t=1}^{T}\pi_\theta(o_t\mid c_t), \\
\pi_T(o\mid x)
&= \prod_{t=1}^{T}\pi_T(o_t\mid c_t).
\end{aligned}
\]
The reverse-KL objective minimized by OPD is
\[
D_{\mathrm{KL}}(\pi_\theta\Vert\pi_T)
=
\mathbb{E}_{x\sim\mathcal{D},\,o\sim\pi_\theta(\cdot\mid x)}
\left[
\log \frac{\pi_\theta(o\mid x)}{\pi_T(o\mid x)}
\right].
\]
Equivalently, OPD maximizes the negative reverse KL:
\[
J_{\mathrm{OPD}}(\theta)
=
\mathbb{E}_{x\sim\mathcal{D},\,o\sim\pi_\theta(\cdot\mid x)}
\left[
\log \frac{\pi_T(o\mid x)}{\pi_\theta(o\mid x)}
\right].
\]
Using the autoregressive factorization, this sequence-level log ratio decomposes into token-level terms:
\[
\log \frac{\pi_T(o\mid x)}{\pi_\theta(o\mid x)}
=
\sum_{t=1}^{T}
\log
\frac{\pi_T(o_t\mid c_t)}
     {\pi_\theta(o_t\mid c_t)}.
\]
This gives the dense token-level OPD reward
\[
r_t^{\mathrm{OPD}}(c_t,o_t)
=
\log
\frac{\pi_T(o_t\mid c_t)}
     {\pi_\theta(o_t\mid c_t)}.
\]

In practice, OPD is optimized with a policy-gradient surrogate in which the reward is treated as a stop-gradient scalar. For a sampled batch of rollout tokens, the empirical objective is
\[
\begin{aligned}
\widehat{J}_{\mathrm{OPD}}(\theta)
=
\sum_{(c_t,o_t)\in\mathcal{B}}
&\mathrm{sg}\!\left[
r_t^{\mathrm{OPD}}(c_t,o_t)
\right] \\
&\cdot \log \pi_\theta(o_t\mid c_t).
\end{aligned}
\]
where \(\mathrm{sg}[\cdot]\) denotes stop-gradient. Taking the gradient gives
\[
\begin{aligned}
\nabla_\theta \widehat{J}_{\mathrm{OPD}}(\theta)
&=
\sum_{(c_t,o_t)\in\mathcal{B}}
\mathrm{sg}\!\left[
r_t^{\mathrm{OPD}}(c_t,o_t)
\right] \\
&\qquad\cdot
\nabla_\theta \log \pi_\theta(o_t\mid c_t).
\end{aligned}
\]
Taking expectation over prompts and student rollouts yields the policy-gradient form
\[
\begin{aligned}
\nabla_\theta J_{\mathrm{OPD}}(\theta)
&=
\mathbb{E}_{\substack{
x\sim\mathcal{D}\\
o\sim\pi_\theta(\cdot\mid x)
}}
\Bigg[
\sum_{t=1}^{T}
\mathrm{sg}\!\left[
\log
\frac{\pi_T(o_t\mid c_t)}
     {\pi_\theta(o_t\mid c_t)}
\right] \\
&\qquad\qquad\cdot
\nabla_\theta \log \pi_\theta(o_t\mid c_t)
\Bigg].
\end{aligned}
\]
Thus, OPD can be implemented as dense-reward policy-gradient learning, where each sampled token receives the stop-gradient reward
\[
r_t^{\mathrm{OPD}}(c_t,o_t)
=
\log
\frac{\pi_T(o_t\mid c_t)}
     {\pi_\theta(o_t\mid c_t)}.
\]
This is the form used throughout the paper.


\end{document}